\documentclass{article}
\usepackage{ijcai20}

\usepackage{times}
\usepackage{soul}
\usepackage{rotating}
\usepackage{url}
\usepackage[hidelinks]{hyperref}
\usepackage[utf8]{inputenc}
\usepackage[small]{caption}
\usepackage{graphicx}
\usepackage{amsmath}
\usepackage{booktabs}
\usepackage[]{mdframed}
\usepackage{algorithm}
\usepackage[]{algcompatible}
\usepackage{algpseudocode}
\usepackage{subcaption}
\usepackage{cite}
\usepackage{amssymb}
\usepackage{multirow}
\usepackage{multicol}
\usepackage{float}

\usepackage{color}
\definecolor{TuBlue}{rgb}{0, 0.65, 0.83}
\definecolor{TuOrange}{rgb}{0.93, 0.41, 0.26}

\newif\ifcomments
\commentsfalse

\newcommand{\algo}[0]{FedDiffuse}

\usepackage{xcolor}
\usepackage{hyperref}
\hypersetup{colorlinks,
citecolor=purple,
linkcolor=black,
urlcolor=purple,
}

\usepackage{flushend}

\title{Training Diffusion Models with Federated Learning
    }

\author{
Matthijs de Goede\and
Bart Cox\and
J\'{e}r\'{e}mie Decouchant\\
\affiliations
Delft University of Technology, The Netherlands\\
\emails
m.b.j.degoede@student.tudelft.nl,
\{b.a.cox, j.decouchant\}@tudelft.nl
}

\begin{document}

\maketitle

\renewcommand{\thepage}{\number\numexpr\value{page}-1\relax}
\pagestyle{plain}
\begin{abstract}

The training of diffusion-based models for image generation is predominantly controlled by a select few Big Tech companies, raising concerns about privacy, copyright, and data authority due to their lack of transparency regarding training data. To address this issue, we propose a federated diffusion model scheme that enables the independent and collaborative training of diffusion models without exposing local data. Our approach adapts the Federated Averaging (FedAvg) algorithm to train a Denoising Diffusion Model (DDPM). Through a novel utilization of the underlying UNet backbone, we achieve a significant reduction of up to 74\% in the number of parameters exchanged during training, compared to the naive FedAvg approach, whilst simultaneously maintaining image quality comparable to the centralized setting, as evaluated by the FID score. Our implementation is publicly available\footnote{\url{https://gitlab.ewi.tudelft.nl/dmls/publications/FedDiffuse}}.

\end{abstract}

\section{Introduction}

Recently, there has been a surge in the popularity of diffusion-based image generation models like Stable Diffusion~\cite{stable_diffusion}, Imagen~\cite{imagegen}, and DALL-E~\cite{dall_e}, which have been praised for their ability to generate synthetic images of exceptional quality and realism. Effective training of these generative models, which typically have hundreds of millions of parameters, requires significant computing power, storage capacities, and a vast amount of training data~\cite{power_concentration}. As a result, most state-of-the-art models are produced by only a handful of Big Tech corporations that have the means to train and maintain them~\cite{power_concentration}. 

Furthermore, the lack of transparency surrounding the origin of the training data of these models raises data authority, privacy, and copyright concerns~\cite{copyright_concerns}. It is often difficult to determine ownership of data obtained from public sources and to ensure informed consent for its use in training machine learning models~\cite{informed_consent}. The inclusion of such data in training processes is problematic as the resulting models may produce outputs that closely resemble copyrighted or sensitive inputs.

To address these issues, we strongly advocate a paradigm shift to a more decentralized approach, where data providers actively participate in training processes, remain in control over their data and consciously share only the strictly required data to produce joint models. This would enable smaller entities and open source communities to participate in the collaborative training of image generation models without compromising their privacy and data authority, thereby decreasing the data and power concentration within Big Tech. A technique that suits this idea is Federated Learning.

\textbf{Federated Learning (FL)}~\cite{federated_original} is a distributed optimization technique that allows multiple clients to collaboratively train a model by leveraging local data. During each training round, a subset of the clients is asked to perform model updates with local data. The local model updates are sent to a central federator server, which performs a global model update based on the aggregated local updates. The updated model is then broadcast to all clients. FL allows for a diverse range of data among clients to be harnessed to build robust models without directly sharing raw data with others, thereby ensuring greater privacy and smaller communication overheads than collaborative methods where raw data is exchanged.

Most of the FL applications today focus on classification and regression tasks. For instance, banks use collaboratively trained models to detect fraudulent transactions~\cite{federated_fraud}, whereas healthcare providers jointly classify sensor data to enhance hospital treatments~\cite{federated_medical}. Federated Learning has also proven to be effective in training large language models across many devices for next-word prediction~\cite{federated_keyboard}.

In the domain of image generation, the use of Federated Learning is still an active research area. Statistical heterogeneity across client datasets and large communication overheads are key challenges in FL~\cite{federated_challenges_methods_directions} that must be overcome to make federated image generation successful. Existing works such as~\cite{fed_gan, fed_gan_2} describe federated techniques based on Generative Adversarial Networks (GANs)~\cite{gan_original}. However, to the best of our knowledge, no federated algorithms have yet been proposed for diffusion models. 

\textbf{Diffusion models} are a type of probabilistic generative models that use noise to gradually destruct training images through multiple forward steps and then learn the reverse denoising process with a neural network to generate new images of the target distribution, given any input of random noise~\cite{diffusion_survey}. Diffusion models are state-of-the-art for image generation as they are more stable in convergence and produce images with higher quality than GANs. However, this comes at the cost of being significantly slower~\cite{diffusion_better}.\\

\noindent This paper aims to bring FL and diffusion models together. More precisely, we address the following research question: \\

\noindent \textit{How can diffusion models for image generation be trained using federated learning?}\\

\noindent To answer this question, we design \algo{}, a Federated Diffusion Model training framework based on a {Denoising Diffusion Probabilistic Model (DDPM)}~\cite{ddpm} that is trained using the Federated Averaging (FedAvg) algorithm~\cite{federated_original}. Additionally, we introduce three novel communication-efficient training methods, \textsc{USplit}, \textsc{ULatDec}, and \textsc{UDec}, that take advantage of the structure of an underlying UNet~\cite{unet} architecture to reduce the number of parameters exchanged during training, whilst maintaining comparable image quality as measured by the FID score~\cite{fid}. 
In a nutshell, \textsc{USplit} splits parameter updates among clients every round, whereas \textsc{ULatDec} and \textsc{UDec} limit the federated training of parameters to specific parts of the network. 
To compare their effectiveness, we evaluate the performance of \algo{} in combination with the different training methods. Finally, we study \algo{} under different data distributions and client settings to assess its robustness to statistical heterogeneity. 

As a summary, we make the following \textbf{contributions}:
\begin{itemize}
    \item We propose a novel algorithm to train diffusion models in a federated way.
    \item We describe and compare three novel communication-efficient training methods that take advantage of the model architecture to reduce the number of communicated parameters during training.
    \textsc{USplit} decreases the communication overhead by 25\%, \textsc{ULatDec} by 41\% , and \textsc{UDec} by 74\%. 
    \item We compare our models by evaluating the image quality of the output images that they generate under different data distributions and client settings. Our results show comparable image quality to the centralized setting in federated settings with up to ten clients and IID data. 
\end{itemize}

\noindent This paper is structured as follows. Section \ref{sec:background} provides background information on federated learning and diffusion models, whereas Section \ref{sec:related_work} sheds light on related research. Section \ref{sec:contributions} explains our communication-efficient methods for federated diffusion, which Section \ref{sec:experiments} tests and compares. Section \ref{sec:conclusions} concludes and provides future work suggestions. 

\section{Background} \label{sec:background}
In this section, we provide the necessary technical background on different types of diffusion models, with a focus on the DDPM. Furthermore, we provide a  formalization of FL and its challenges with statistical heterogeneity. \\

\textbf{Types of Diffusion Models.}
Among diffusion models, we distinguish between three predominant formulations. First, Denoising Diffusion Probabilistic Models (DDPMs)~\cite{diffusion_original, ddpm} estimate a probability distribution over image data using a diffusion process over discrete timesteps, with both forward and reverse processes represented as Markov chains. Second, Score-based Generative Models (SGMs)~\cite{score_based_diffusion, improved_score_based_diffusion} learn the Stein Score~\cite{stein_score}, which represents the gradient of the log-density function of the image data. During sampling, noisy inputs pass discrete timesteps in the reverse process at which they are pushed in the direction in which the data density, and thus sample likelihood grows the most. Third, Stochastic Differential Equations (Score SDEs)~\cite{stochastic_differential_equations} are the continuous-time generalization of both SGMs and DDPMs that estimate the score function at any time using differential equations.

We choose to focus on the DDPM formulation, mainly because of its simplicity and popularity. The loss-based objective function is easier to optimize than the score-based objectives that SGMs and SDEs use. Once the transition kernels are learned, no numerical methods are required to generate samples, unlike with SDEs. The DDPM is also the most explored and widespread option out of the three~\cite{diffusion_survey}. \\

\textbf{Denoising Diffusion Models (DDPM).}
The DDPM introduced by~\cite{ddpm} models a probability distribution $p_\theta(x_0) := \int p_\theta(x_{0:T})dx_{1:T}$, over the pixel space through noisy latents $x_1,...,x_T$.  Given training images $x_0$ from a noiseless target distribution $q(x_0)$, the latents $x_1,...,x_T$ are obtained following a Markovian forward process $q(x_{1:T})$ that gradually adds Gaussian noise according to a variance schedule $\beta_{1},...,\beta_{T}$, as given by equations \ref{eq:1} and \ref{eq:2}.
\begin{align}
    q(x_{1:T}) &:= \prod_{t=1}^{T}{q(x_t|x_{t-1})} \label{eq:1} \\
    q(x_t|x_{t-1}) &:= \mathcal{N}(x_t;\sqrt{1-\beta_t}x_{t-1}, \beta_t I) \label{eq:2}
\end{align}

\noindent Provided that the variance schedule is chosen so that $\bar{\alpha}_{T} = \prod_{s=1}^{T}({1-\beta_{s}}) \to 0$, the distribution of $x_T$ is well approximated by the standard Gaussian (random noise) distribution $p(x_T) \approx \mathcal{N}(x_T;0, I)$~\cite{diffusion_survey}. In the reverse process, the goal is to create a noiseless sample starting with a sample of random noise. When the $\beta_t$ are sufficiently small, the reverse process has the same functional form as the forward process. Therefore, the reverse process can be defined by a Markov chain $p_{\theta}(x_{0:T})$ with learned Gaussian transitions parameterized by $\theta$, as given by Equations \ref{eq:3} and \ref{eq:4} below.    
\begin{align}
 p_{\theta}(x_{0:T}) &:= p(x_T)\prod_{t=1}^{T}{p_{\theta}(x_{t-1}|x_{t})} \label{eq:3}\\
p_{\theta}(x_{t-1}|x_{t}) &:= \mathcal{N}(x_{t-1}; \mu_\theta(x_t, t), \Sigma_{\theta}(x_t, t)) \label{eq:4}
\end{align}

\noindent In~\cite{ddpm}, the variances of the denoising kernels are fixed to a single value: $\Sigma_{\theta}(x_t, t) = \sigma_{t}^2 I$, where $\sigma_{t}^2 = \frac{1-\bar{\alpha}_{t-1}}{1-\bar{\alpha}_{t}}\beta_{t}$ and $\bar{\alpha}_{t} = \prod_{s=1}^{t}({1-\beta_{s}})$. However, they can also be learned during training~\cite{improved_ddpm}. 
 Instead of approximating $\mu_\theta(x_t, t)$ directly, it is re-parameterized as a function of $\epsilon_{\theta}(x_t, t)$ to achieve better sampling quality~\cite{ddpm}. $\epsilon_{\theta}(x_t, t)$ approximates the noise $\epsilon_t$ that is to be subtracted from samples $x_t$ at timestep $t$ during the reverse process:

 %comment
 %comment

 \begin{align}
     \mu_{\theta}(x_t, t) &= \frac{1}{\sqrt{1-\beta_t}}(x_t-\frac{\beta_t}{\sqrt{1-\bar{\alpha}_{t}}}\epsilon_{\theta}(x_t, t))
 \end{align}

\noindent A special property of the forward process is that:

 \begin{align}
     q(x_t|x_0) &:= \mathcal{N}(x_t; \sqrt{\bar{\alpha}_t}x_0, (1-\bar{\alpha}_t)I)
 \end{align}

 \noindent Using this, any noisy latent $x_t$ can be sampled via a single step given the original image $x_0$ and fixed variances $\beta_t$:  

 \begin{align} \label{eq:x_t}
     x_t = \sqrt{\bar{\alpha}_t}x_0 + \sqrt{1 - \bar{\alpha}_t} \epsilon_t \quad \text{where} \quad \epsilon_t \sim \mathcal{N}(0, I) 
 \end{align}

 \noindent The training objective can be formulated as minimizing the distance between the real noise $\epsilon_t$ and the noise estimation $\epsilon_{\theta}(x_t, t)$ by the model for each of the timesteps $t$: 

 \begin{align} \label{eq:loss}
     \mathcal{L}_{\text{simple}}(\theta) &:= \mathbb{E}_{t\sim[1,T]}\mathbb{E}_{x_0\sim p(x_0)}\mathbb{E}_{\epsilon_t\sim N(0,I)}\|\epsilon_t - \epsilon_{\theta}(x_t, t)\|_2^2
 \end{align}

\noindent Here, $\mathcal{L}_{simple}$ is a simplified objective function derived from the variational lower bound on the negative log-likelihood for parameter $\theta$ ($\mathcal{L}_{vlb}$)\cite{ddpm}. 
We can learn $\theta$ by using a neural network trained on minimizing $\mathcal{L}_{\text{simple}}$  using Stochastic Gradient Descent (SGD), as shown in Algorithm \ref{algo:training}.

\begin{algorithm}[H]
\caption{DDPM Training Algorithm}
\label{algo:training}
\begin{algorithmic}[0]
\State \textbf{repeat}
\State \quad $x_0 \sim q(x_0)$ 
\State \quad $t \sim Uniform(\{1,...,T\})$ 
\State \quad $\epsilon_t \sim \mathcal{N}(0, I)$
\State \quad \text{Take a gradient descent step on} \\  $ \quad \quad \nabla_{\theta}{\|\epsilon_t - \epsilon_{\theta}(\sqrt{\bar{\alpha}_t}x_0 + \sqrt{1-\bar{\alpha}_t}\epsilon_t, t)\|_2^2}$
\State \textbf{until} converged
\end{algorithmic}
\end{algorithm} 

\noindent Once trained, a DDPM can generate images via Algorithm \ref{algo:sampling}.

\begin{algorithm}[H]
\caption{DDPM Sampling Algorithm}
\label{algo:sampling}
\begin{algorithmic}[0]
\State $x_T \sim \mathcal{N}(0, I)$
\For{$t = T$ \textbf{down to} $1$}
\State $z \sim \mathcal{N}(0, I)$ if $t > 1$, else $z = 0$
\State $x_{t-1} = \mu_{\theta}(x_t, t) + \sigma_tz$

\EndFor
\State \textbf{return} $x_0$
\end{algorithmic}
\end{algorithm}

\begin{figure}[h]
\includegraphics[width=\linewidth]{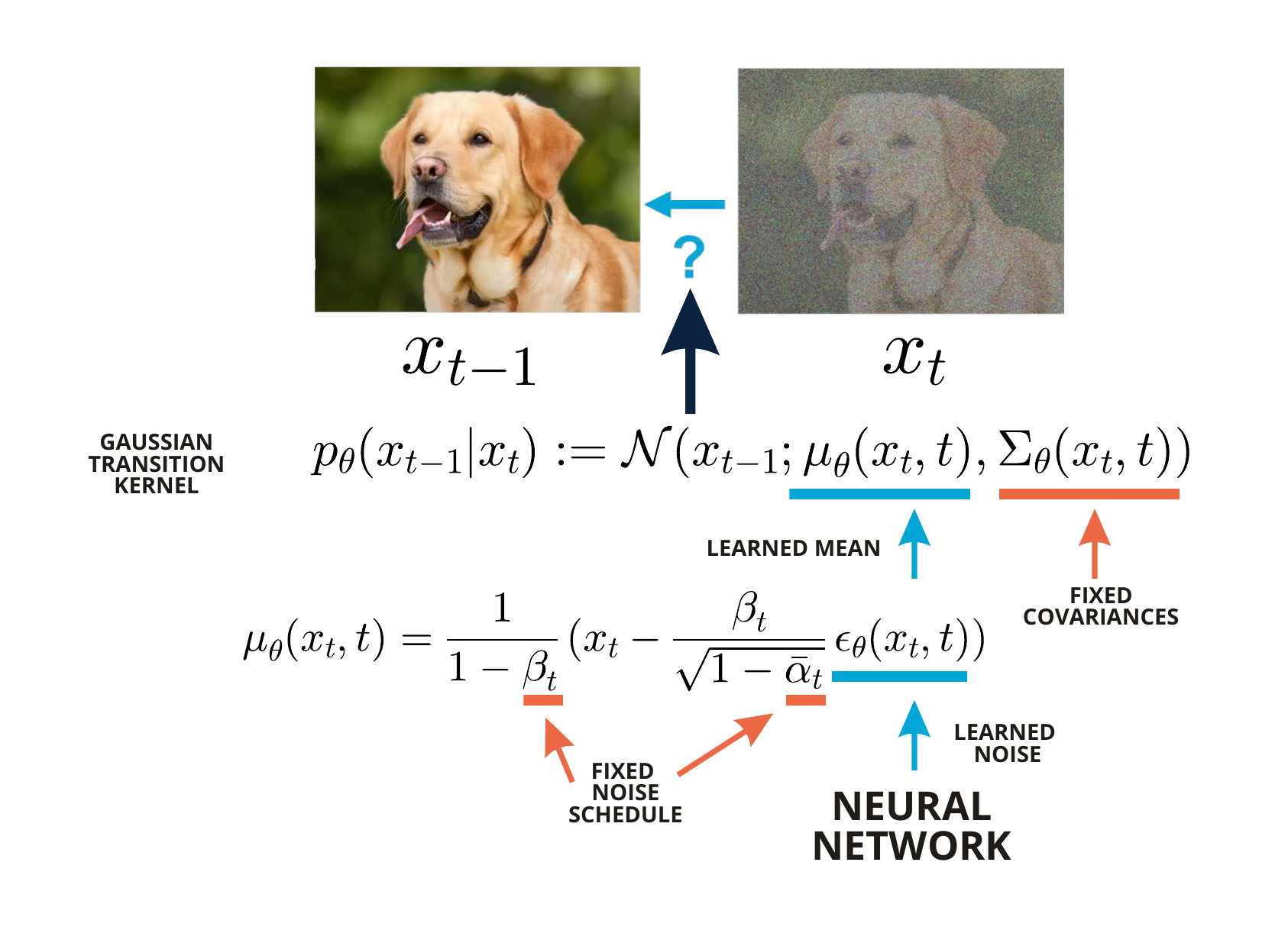} 
\caption{Graphical representation of the intuition behind the DDPM. The reverse denoising process uses Gaussian transition kernels with fixed covariances $\Sigma_{\theta}(x_t, t)$ and means $\mu_\theta(x_t,t)$ that are learned using a neural network predicting the noise $\epsilon_\theta(x_t,t)$ to subtract from samples $x_t$ at each timestep $t$.}
\label{fig:ddpm}
\end{figure}

\noindent Figure \ref{fig:ddpm} provides the intuition behind the DDPM model. \\

\textbf{Formalization of Federated Learning.} 
A typical federated learning problem can be formulated as a distributed optimization problem involving $K$ clients that aim at minimizing the following objective function~\cite{federated_original}: 

\begin{align} \label{eq:fl}
    \min_{\theta \in \mathbb{R}^d} f(\theta), \quad \text{where} \quad f(\theta) &:= \frac{1}{K}\sum_{k=1}^{K} w_k f_k(\theta)
\end{align}

\noindent For a deep learning problem, $f_k(\theta)$ typically represents the loss incurred over a local client dataset $D_k \subset D$  under global model parameter vector $\theta$. The impact $w_k$ that a client $k$ has on the global objective is often weighted by the relative size of its dataset so that $w_k = \frac{|D_k|}{|D|}$.  \\

\textbf{Statistical Heterogeneity in Federated Learning.} \label{sec:stat_het}If the client datasets $D_k$ are formed by distributing the training examples over the $K$ clients uniformly at random, the data is said to be Independent and Identically Distributed (IID). In this case, we have that $\mathbb{E}_{D_k}[f_k(\theta)] = f(\theta)$ for all clients~\cite{federated_original}. Cases where this does not hold are referred to as statistically heterogeneous or non-IID. In such cases, there is no guarantee that the $f_k$ estimate $f$ well. Dealing with statistical heterogeneity is one of the main challenges within FL~\cite{federated_advances, federated_challenges_methods_directions,cox2022aergia}. 

Considering the federated diffusion scenario, we focus on two causes for statistical heterogeneity. First, there can be significant differences in the number of training images each client contributed, referred to as \textbf{quantity skew}. To address this, client model updates can be weighted based on their respective dataset sizes~\cite{non_iid}. Second, in the context of labeled datasets, image label distributions may vary among clients, which is known as \textbf{label distribution skew}. Handling label distribution skew can be challenging because each client tends to adjust its local model toward its most dominant labels,
resulting in different update directions that need to be combined~\cite{non_iid}. 

\section{Related Work} \label{sec:related_work}

To the best of our knowledge, no previous work has attempted to combine diffusion models and FL. However, a federated algorithm to train image segmentation models with a similar architecture as diffusion models has been proposed by Kanagavelu et al.~\cite{fed_ukd}. Moreover, notable works have explored alternative solutions for federated image generation based on GANs~\cite{fed_gan, fed_gan_2}. Additionally, numerous papers focused on enhancing communication efficiency within the context of FL~\cite{comm_eff_google, dist_mean_est, gradient_quant}. Finally, it is worth mentioning Latent Diffusion Models (LDMs)~\cite{stable_diffusion}, which perform the diffusion process in a low dimensional latent space, resulting in fewer parameters to optimize and exchange.  \\

\textbf{Federated UNet.} The transition kernels for the reverse process of diffusion models are usually learned using architectures that build upon the UNet~\cite{unet} convolutional network~\cite{ddpm, improved_ddpm, diffusion_better, stable_diffusion}. As the UNet model was initially developed for image segmentation, it is no surprise that the first federated solution centers around this task. Namely,~\cite{fed_ukd} introduces a federated UNet model to segment satellite images based on land use. Aggregation of the local model updates at the federator is performed using FedAvg~\cite{federated_original}. The model is shown to perform well on label-skewed datasets. However, the used datasets contain few images, which is typical for image segmentation problems but differs from the image generation scenario. The authors further claim spectacular compression rates for both the number of parameters as well as the memory taken by these parameters, although no further details are provided. \\

\textbf{Federated Image Generation.} Generative Adversarial Networks (GANs)~\cite{gan_original} used to dominate the field of image generation before diffusion models surpassed them in terms of image fidelity and training stability~\cite{diffusion_better}. GANs differ from diffusion models in terms of their architectural approach. Diffusion models utilize a single network to make noise predictions at each timestep of the denoising process, whereas GANs employ two networks: a generator that directly generates output images from noise, and a discriminator that classifies the produced images as real or fake to steer the generator.

GANs have a rich research history that also includes the cross-silo~\cite{federated_advances} federated setting. Specifically,~\cite{fed_gan_2} proposed a federated GAN framework and tested different synchronization strategies with up to six clients to determine whether training either the generator or discriminator collaboratively whilst training the other locally would yield comparable results to training both components in a federated manner, which was found not to be the case. Additionally, the study revealed that federated training of GANs becomes less effective when the data distribution is more skewed and that this effect becomes more pronounced as the number of clients increases. We pose a similar hypothesis for federated diffusion. 

Alternatively,~\cite{fed_gan} proposes a communication-efficient method where the discriminator and generator are trained by averaging over the local parameter values only every $K$ rounds. They show that the model's performance is robust to increasing the synchronization interval K, in a setting with five clients. Additionally, they provide a formal proof on the convergence of the algorithm in non-IID scenarios. \\

\textbf{Improving Communication Efficiency.} Various works have looked into compression and quantization methods to reduce message size in FL~\cite{comm_eff_google, dist_mean_est, gradient_quant}. With stochastic $k$-level quantization, a limited number of $\log_2 k$ bits is used to represent each of the coordinates within a gradient vector. Each coordinate is rounded to one of the $k$ evenly spread levels between the minimum and maximum value of the corresponding coordinate. Variable length encodings for each of the coordinates can subsequently be applied to further reduce the number of bits transmitted to the federator~\cite{dist_mean_est}.

Alternative methods include gradient sparsification, where only a subset of the gradients is sent to the federator based on absolute values, thresholds, or random bitmasks, and low-rank decomposition, where a model update is represented as the product of two low-rank matrices, out of which only one is trained and sent to the federator, whilst the other is initialized randomly every round~\cite{comm_eff_google}.

More recently,~\cite{correlated_quantization} introduced correlated quantization, which uses shared randomness to introduce correlation between the local quantizers at each client, improving error bounds and speeding up convergence. The main intuition behind correlated quantization is that if the first client rounds up its value, the second client should round down its value to reduce the mean squared error. 

The research on compression and quantization methods is mainly based on general statistical methods that could also be applied to diffusion gradients. However, none of the methods seems to take advantage of the underlying model architecture, so that we consider them orthogonal to our work.  \\

\textbf{Latent Diffusion Models.} A recent breakthrough in diffusion research is the Latent Diffusion Model (LDM)~\cite{stable_diffusion}, where the diffusion process takes place in a latent space of reduced dimensionality rather than the high dimensional RGB picture space. It was found that most of the bits from input images relate to perceptual rather than semantic or conceptual composition so that the images could aggressively be compressed without losing information about the latter. A major benefit of this approach is the reduced number of parameters to be optimized in the UNet~\cite{unet} to approximate the denoising process. This is especially fruitful in a federated setting where the weights have to be sent back and forth between clients and the federator. A downside of this approach is that it requires a separately trained encoder and decoder to convert between the image and latent space. 

\section{Communication Efficient Federated Diffusion} \label{sec:contributions}

In this section, we explain our federated diffusion algorithm \algo{} as well as the underlying UNet architecture and our communication efficient training methods, \textsc{USplit}, \textsc{UDec}, and \textsc{ULatDec}, which take advantage of this architecture. \\

\textbf{Federated Diffusion.} In our federated diffusion scenario, we consider a cross-silo setting \cite{federated_advances} with a small set of $K$ clients equipped with reasonable computing power and relatively large datasets $D_k \in D$. We use the Federated Averaging (FedAvg) algorithm \cite{federated_original} to optimize the objective from Equation \ref{eq:fl}, as it has proven to be capable of training a wide variety of deep neural networks using relatively few rounds of communication between the federator and the clients.

Initially, we randomly initialize a global model with parameter vector $\theta_0$. We introduce $R$ training rounds in which all clients partake. They receive the latest model parameters $\theta_{r-1}$ from the federator at the start of each round $r$ and perform SGD minimizing $\mathcal{L}_{\text{simple}}$ over their local dataset $D_k$ to produce an updated parameter vector $\theta_r^{k}$, such as in Algorithm \ref{algo:training}. Specifically, we use mini-batch SGD with batch size $B$, and fixed learning rate $\eta$.  Parameter $E$ regulates the number of local epochs that every client performs over its dataset every round. At the end of every round, the clients send back $\theta_r^{k}$ to the federator, which takes a weighted sum over the client vectors using the relative dataset size $\frac{|D_k|}{|D|}$ to produce an updated global model with parameters $\theta_r$.  Algorithm \ref{alg:fed_diff} details \algo{}'s pseudocode.

\begin{algorithm}[H]
\caption{Federated Diffusion (\algo{})}
\label{alg:fed_diff}
\begin{algorithmic}[0]
\State \textbf{Input}: Number of clients $K$, number of communication rounds $R$, number of local epochs $E$, local mini-batch size $B$, local datasets $D^k$, learning rate $\eta$, number of diffusion timesteps T and variance schedule $\beta_{1},...,\beta_{T}$.
\State \textbf{Output}: Global model parameters $\theta_R$\\

\State \textbf{Federator executes}: 
\State initialize $\theta_0$
\State $|D| \gets \sum_{k=1}^{K}{|D^k|}$ 

\For {$r = 1$ to $R$}

    \For {$k = 1$ to $K$}
        \State $\theta_{r}^{k} \gets \textproc{ClientUpdate}(k, \theta_{r-1})$
    \EndFor
    
    \State $\theta_r \gets \frac{1}{|D|}\sum_{k=1}^{K}\theta_{r}^{k} \cdot |D^k|$
\EndFor
\State
\State \textbf{Client executes}:
\Function{ClientUpdate}{$k, \theta_{r-1}$}:

    \State $\theta_r^k \gets \theta_{r-1}$    
    \State $\mathcal{B} \gets$ (split $D^k$ into batches of size $B$)

    \For {$e = 1$ to $E$}
        \For {$b \in \mathcal{B}$} 
            \State $\theta_r^k \gets \theta_r^k - \eta \cdot \nabla_{\theta_r^{k}} \textproc{CalculateLoss}(b;\theta_r^k) 
            $
        \EndFor
    \EndFor
    \State \Return $\theta_{r}^{k}$
\EndFunction
\State
\Function{CalculateLoss}{$b;\theta_r^k$}:
    \For {$i \in b$}
        \State $t \sim Uniform(\{1,..,T\})$
        \State $\epsilon_t \sim \mathcal{N}(0, I)$
        \State $\bar{\alpha}_{t} = \prod_{s=1}^{t}({1-\beta_{t}})$
        \State $\mathcal{L}_i = {\|\epsilon_t - \epsilon_{\theta_r^k}(\sqrt{\bar{\alpha}_t}i + \sqrt{1-\bar{\alpha}_t}\epsilon_t, t)\|_2^2}$
    \EndFor
    \State \Return $\frac{1}{|b|}\sum_{i \in b} \mathcal{L}_i$
\EndFunction
\end{algorithmic}
\end{algorithm}

\textbf{UNet Architecture.} \label{sec:unet} For every client, we use an identical UNet \cite{unet} convolutional neural network to approximate the function $\epsilon_{\theta}(x_t, t)$. The name "UNet" is derived from the network's U-shaped architecture, which consists of an encoder and a decoder path with what is referred to as a latent bridge or bottleneck in the middle. First, the encoder path gradually downsamples the noisy input images to capture an increasing number of higher-level but lower-resolution feature maps. The bottleneck in the middle can then be leveraged to perform feature selection, after which the decoder path performs upsampling to generate pixel-level predictions of the noise $\epsilon_t$.  Skip connections inspired by \cite{resnet} are employed to bridge the gap between the encoder and decoder, allowing the network to combine both low-level and high-level features effectively.

\begin{figure*}[!t]
\includegraphics[width=\linewidth]{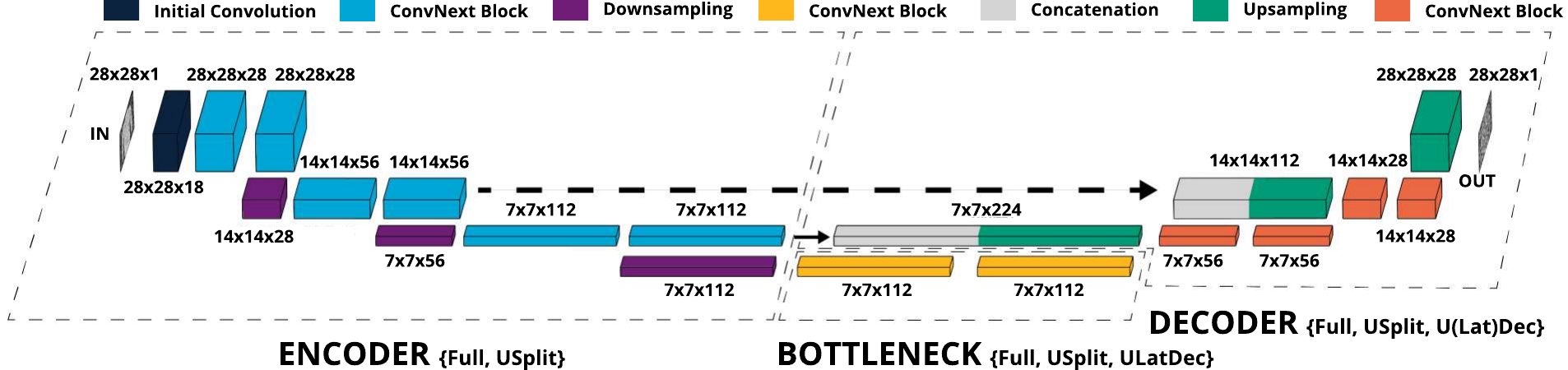}
  \centering
  \caption{UNet depiction showing the widths, heights, and counts for the feature maps resulting from the different operations in the encoder, bottleneck, and decoder. For each network part, the training methods that consider it for federated training are indicated within the brackets.}
  \label{fig:unet}
\end{figure*}

In our version, the Wide ResNet Blocks \cite{resnet} used by \cite{ddpm} are replaced by more state-of-the-art ConvNeXt Blocks \cite{convnet}. Another difference is that we apply three rather than four levels of downsampling because we aim at generating small 28x28 images. Our bottleneck preserves spatial dimensionality and feature map count to allow a smooth gradient flow between the encoder and decoder and straightforward concatenation via the skip connections in the layers above. Parameter sharing over time is accommodated by leveraging transformer sinusoidal position embeddings \cite{attention} for the diffusion timesteps $t$, as in \cite{annotated_diffusion}. A graphical representation of our UNet model, showing the feature map dimensions and counts resulting from the operations in the encoder, bottleneck, and decoder can be found in Figure \ref{fig:unet}. \\

\textbf{Communication Efficient Training Methods.} By default \algo{} uses what we refer to as the \textsc{Full} training method, which consists of the federator sending the full parameter vector $\theta$ to each of the $K$ clients and receiving the updated parameter vectors $\theta^k$ from each of the clients during each of the $R$ communication rounds. Let $\theta_{\text{enc}}$, $\theta_{\text{bot}}$, $\theta_{\text{dec}}$ be the parameter vectors associated with the UNets encoder, bottleneck and decoder respectively so that $\theta = \theta_{\text{enc}} \frown \theta_{\text{bot}} \frown \theta_{\text{dec}}$, where the $\frown$ operator denotes vector concatenation. The total communication overhead of \textsc{Full} is now $\mathcal{O}(R \cdot K \cdot 2|\theta|)$.

\noindent We propose two alternative types of training techniques that exploit the structure of the UNet to reduce the total communication overhead incurred during the training process. \\

\textbf{\textsc{USplit}} decreases the communication overhead by splitting parameter updates complementarily amongst the clients. The federator initiates each communication round again by sending the full parameter vector $\theta$ to each of the clients so that these can initialize their local model identically. 
However, each client is assigned a specific subset of the parameters, which can include $\theta_{\text{enc}}$, $\theta_{\text{bot}}$ and/or $\theta_{\text{dec}}$, to report the updates for that round. The global model is then updated using an adapted version of \algo{} that only considers the updates from the responsible clients for each network part. 

In more detail, tasks are assigned as follows:
Every round, we divide the set of clients into random pairs. In each pair, one client reports about the encoder and the other about the decoder. The task of reporting about the bottleneck is randomly assigned to one of the two. If the number of clients is odd, the last client is assigned either the encoder or decoder task randomly, in addition to the bottleneck task. 

This task assignment method mimics selecting a random fraction $C = 0.5$ of the clients every round to perform model updates, like in \cite{federated_original}. However, this is now done for each of the network parts independently. By assigning new tasks every round, the federator still gathers information regarding each of the network parts for each of the clients over time, whilst reducing the communication overhead of the client updates by a factor of two. As the communication overhead introduced by the federator remains the same, this results in an overall overhead in $\mathcal{O}(R \cdot K \cdot \frac{3}{2}|\theta|)$.

Alternatively, \textbf{\textsc{UDec}} and \textbf{\textsc{ULatDec}} limit the federated training of the model to a subset of the parameters, and leave the training of the other parameters up to the clients themselves. This results in every client having a composed model with both globally trained as well as locally trained parameters, much like in Transfer Learning \cite{transfer_learning}.

The intuition behind both methods is that the denoising capacity of the UNet can mainly be attributed to the decoder, which creates the noise estimations based on the features extracted and selected by the encoder and bottleneck respectively. Hence, \textsc{UDec} collaboratively trains (and thus exchanges) only the decoder parameters. 
As a result, clients have the freedom to utilize their locally trained encoder and bottleneck to extract and select features. This might result in mismatches between the locally selected features and the features expected as inputs to the decoder. \textsc{ULatDec} aims to mitigate this issue by training the bottleneck collaboratively too, so that the feature selection is more unified. As the bottleneck in our UNet does not perform explicit feature selection by reducing the number of feature maps, we expect little difference in model performance between both methods. \textsc{UDec} and \textsc{ULatDec} have a communication overhead of $\mathcal{O}(R \cdot K \cdot 2|\theta_{\text{dec}}|)$ and $\mathcal{O}(R \cdot K \cdot 2|\theta_{\text{dec}} \frown \theta_{\text{bot}}|)$ respectively.

\section{Experimental Setup and Results} \label{sec:experiments}

In this section, we first describe our experimental setup and evaluation metrics. Then we describe the different experiments that we carried out to quantitatively evaluate our methods in different federated settings and discuss their results. \\

\textbf{Experimental Details.} All models were implemented using the PyTorch framework. We used the Fashion-MNIST dataset \cite{fmnist}, which consists of 60,000 training and 10,000 test images of 10 different fashion items in grayscale, each having 28x28 pixels. The diffusion parameters from \cite{ddpm} were adopted, specifically $T = 1000$ and the linear diffusion schedule ranging from $\beta_1 = 10^{-4}$ to $\beta_T = 0.02$. Our model of choice was the UNet, as discussed in Section \ref{sec:unet}, which contained a total of 2,996,315 parameters. For the SGD optimizer, we used local batch size $B = 128$ and learning rate $\eta = 10^{-4}$. To damp out gradient oscillations, we employed the Adam optimizer \cite{adam}. All experiments were conducted on a single NVIDIA GeForce RTX 3090 GPU with CUDA 11.7. We performed 5 runs per experiment and reported averages. \\

\textbf{Evaluation Metrics.} To evaluate the communication efficiency of our models, we reported the cumulative number of communicated parameters between the federator and all clients during model training ($N$). To measure image quality, we used the widespread Fréchet Inception Distance (FID) \cite{fid}, which measures the distance between a target distribution and a distribution of generated samples based on mean vectors and covariance matrices extracted by a pre-trained Inception V3 model \cite{inception}. The lower the FID, the better the image quality. Usually, 50.000 images per distribution are used to extract the required statistics, but given the slow diffusion sampling and the fact that our global test set only contained 10,000 images, we decided to use 5,000 images instead. We measured the FIDs on client level, given that the federator only had access to partial models with \textsc{ULatDec} and \textsc{UDec}. \\

\textbf{Establishing a Centralized Baseline.} We first considered the centralized setting where $K=1$ and trained models with $R=30$. We visually estimated the quality of the output images and found this to be sufficient after 10 rounds of training. Hence, we set the corresponding mean FID of 72 as the image artifact threshold, below which quality was deemed acceptable. We further established that there was little improvement from round 15 onwards. Hence, we set the corresponding mean FID of 43 as the centralized baseline and fixed $R=15$ for the federated setting to compare with. \\

\begin{figure}[t]
\includegraphics[width=\linewidth]{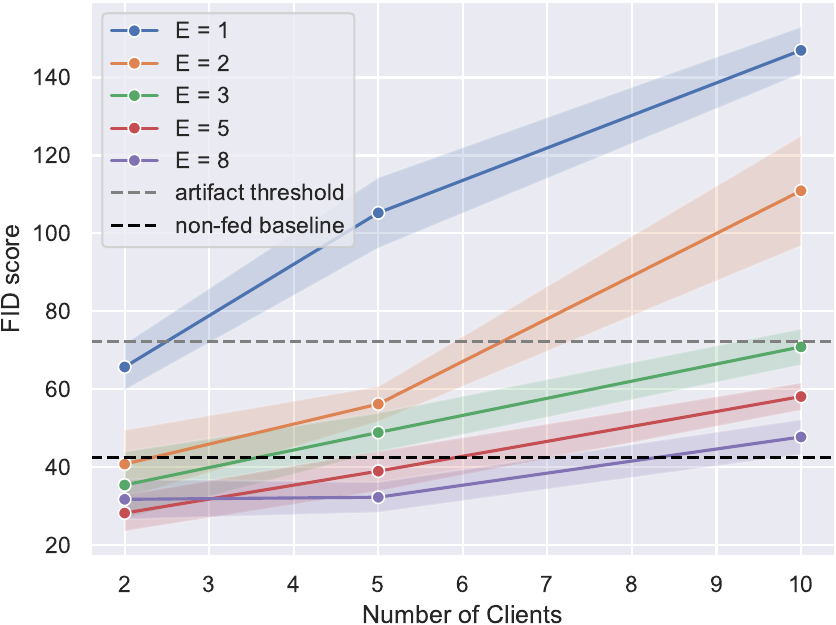} 
\caption{Mean FID scores with error bounds for different number of clients $K$ and local epochs $E$ with $R=15$ in the \textsc{Full} federated setting on IID data.}
\label{fig:num_clients}
\end{figure}

\textbf{Testing the Federated Setting.} Next, we conducted experiments in the \textsc{Full} federated setting, testing different numbers of clients $K \in \{2,5,10\}$ on IID data using $R=15$ and $E=1$. Figure \ref{fig:num_clients} demonstrates that the FID scores quickly surpassed the artifact threshold as the number of clients increased. To achieve better FID scores without increasing the number of communication rounds, we explored different numbers of local epochs $E \in \{2,3,5,8\}$ per communication round. As shown in Figure \ref{fig:num_clients}, increasing $E$ significantly improved the FID scores. The higher the number of clients $K$, the more local epochs $E$ were required to bring the FID scores under the artifact threshold. However, the training time linearly increased with $E$. To strike a balance between training time and output quality, we opted for $E=5$, which yielded FID scores that were comparable with the centralized baseline, whilst maintaining reasonable maximum training times at around 30 minutes per model. \\

\textbf{Comparison of the Training Methods.} With the number of epochs $E=5$ and global communication rounds $R=15$ fixed, we compared the \textsc{Full} federated training with \textsc{USplit}, \textsc{ULatDec} and \textsc{UDec} in terms of the cumulative number of communicated parameters $N$ and the resulting FIDs for different number of clients $K \in \{2,5,10\}$ with IID data.

\begin{figure}[htbp]
\includegraphics[width=\linewidth]{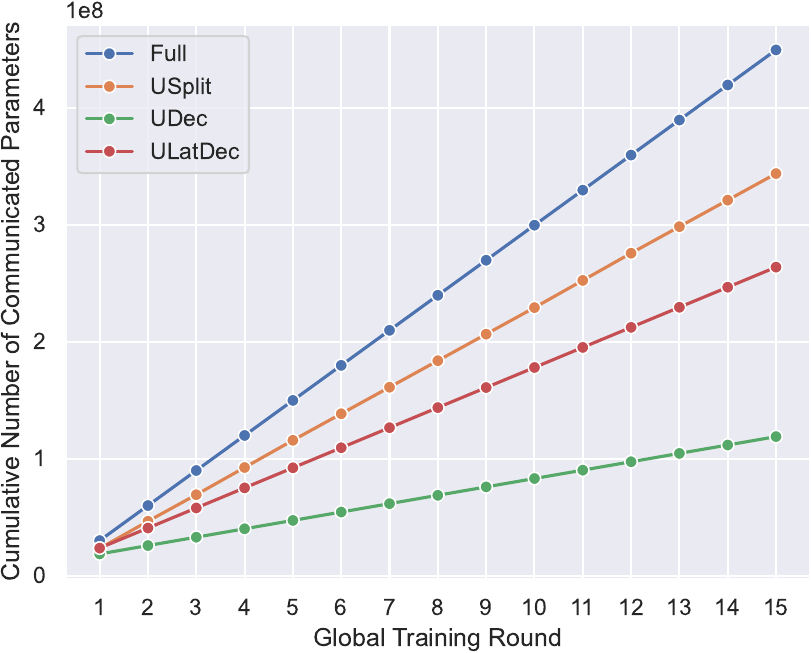} 
\caption{Cumulative number of communicated parameters $(\cdot 10^8)$ during training for the different training methods with $K=5$.}
\label{fig:num_params}
\end{figure}

Figure \ref{fig:num_params} shows the linear development of $N$ over the training rounds for each of the methods with $K=5$, whereas Table \ref{tab:results} shows $N$ for each of the settings. On average, \textsc{USplit} achieved a 25\% reduction over \textsc{Full},  where \textsc{ULatDec} and \textsc{UDec} achieved a 41\% and 74\% reduction respectively. These are in correspondence with the Big-$\mathcal{O}$ bounds for communication overhead established in Section \ref{sec:contributions}. 

\noindent Table \ref{tab:results} shows comparable FID scores for \textsc{USplit} and \textsc{Full} in the IID setting, where \textsc{UDec} and \textsc{ULatDec} have higher FID scores. There is little difference between the latter two, which is in line with our hypothesis that training the latent bridge in a federated manner would not significantly improve the image quality for our version of the UNet. In future work, we plan to explore different bottleneck configurations to investigate their effect on both training methods.

Another noteworthy observation concerns the higher standard deviations for \textsc{UDec} and \textsc{ULatDec}, in comparison to \textsc{Full} and \textsc{USplit}. These can be attributed to performance variations across local client models resulting from partial federated training, as elucidated in Table \ref{tab:client_fids}. For instance, the FID scores of Client 3 are twice as high as those of Client 1, indicating that Client 1 was strikingly more successful in training the encoder and bottleneck locally than Client 3, even though their training data was IID. 

\begin{figure}[H]
  \centering
  
  \begin{subfigure}{0.09\linewidth}
    \includegraphics[width=\linewidth]{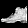}
  \end{subfigure}
  \begin{subfigure}{0.09\linewidth}
    \includegraphics[width=\linewidth]{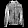}
  \end{subfigure}
  \begin{subfigure}{0.09\linewidth}
    \includegraphics[width=\linewidth]{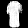}
  \end{subfigure}
  \begin{subfigure}{0.09\linewidth}
    \includegraphics[width=\linewidth]{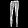}
  \end{subfigure}
  \begin{subfigure}{0.09\linewidth}
    \includegraphics[width=\linewidth]{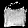}
  \end{subfigure}
  \begin{subfigure}{0.09\linewidth}
    \includegraphics[width=\linewidth]{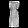}
  \end{subfigure}
  \begin{subfigure}{0.09\linewidth}
    \includegraphics[width=\linewidth]{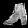}
  \end{subfigure}
  \begin{subfigure}{0.09\linewidth}
    \includegraphics[width=\linewidth]{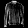}
  \end{subfigure}
  \begin{subfigure}{0.09\linewidth}
    \includegraphics[width=\linewidth]{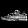}
  \end{subfigure}
  \begin{subfigure}{0.09\linewidth}
    \includegraphics[width=\linewidth]{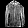}
  \end{subfigure}

  \vspace{0.2em}

  \begin{subfigure}{0.09\linewidth}
    \includegraphics[width=\linewidth]{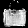}
  \end{subfigure}
  \begin{subfigure}{0.09\linewidth}
    \includegraphics[width=\linewidth]{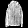}
  \end{subfigure}
  \begin{subfigure}{0.09\linewidth}
    \includegraphics[width=\linewidth]{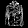}
  \end{subfigure}
  \begin{subfigure}{0.09\linewidth}
    \includegraphics[width=\linewidth]{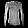}
  \end{subfigure}
  \begin{subfigure}{0.09\linewidth}
    \includegraphics[width=\linewidth]{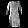}
  \end{subfigure}
  \begin{subfigure}{0.09\linewidth}
    \includegraphics[width=\linewidth]{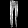}
  \end{subfigure}
  \begin{subfigure}{0.09\linewidth}
    \includegraphics[width=\linewidth]{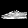}
  \end{subfigure}
  \begin{subfigure}{0.09\linewidth}
    \includegraphics[width=\linewidth]{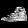}
  \end{subfigure}
  \begin{subfigure}{0.09\linewidth}
    \includegraphics[width=\linewidth]{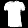}
  \end{subfigure}
  \begin{subfigure}{0.09\linewidth}
    \includegraphics[width=\linewidth]{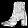}
  \end{subfigure}

  \vspace{0.2em}

    \begin{subfigure}{0.09\linewidth}
    \includegraphics[width=\linewidth]{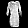}
  \end{subfigure}
  \begin{subfigure}{0.09\linewidth}
    \includegraphics[width=\linewidth]{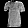}
  \end{subfigure}
  \begin{subfigure}{0.09\linewidth}
    \includegraphics[width=\linewidth]{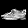}
  \end{subfigure}
  \begin{subfigure}{0.09\linewidth}
    \includegraphics[width=\linewidth]{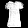}
  \end{subfigure}
  \begin{subfigure}{0.09\linewidth}
    \includegraphics[width=\linewidth]{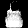}
  \end{subfigure}
  \begin{subfigure}{0.09\linewidth}
    \includegraphics[width=\linewidth]{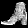}
  \end{subfigure}
  \begin{subfigure}{0.09\linewidth}
    \includegraphics[width=\linewidth]{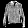}
  \end{subfigure}
  \begin{subfigure}{0.09\linewidth}
    \includegraphics[width=\linewidth]{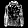}
  \end{subfigure}
  \begin{subfigure}{0.09\linewidth}
    \includegraphics[width=\linewidth]{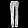}
  \end{subfigure}
  \begin{subfigure}{0.09\linewidth}
    \includegraphics[width=\linewidth]{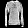}
  \end{subfigure}

  \vspace{0.2em}

    \begin{subfigure}{0.09\linewidth}
    \includegraphics[width=\linewidth]{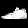}
  \end{subfigure}
  \begin{subfigure}{0.09\linewidth}
    \includegraphics[width=\linewidth]{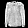}
  \end{subfigure}
  \begin{subfigure}{0.09\linewidth}
    \includegraphics[width=\linewidth]{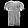}
  \end{subfigure}
  \begin{subfigure}{0.09\linewidth}
    \includegraphics[width=\linewidth]{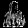}
  \end{subfigure}
  \begin{subfigure}{0.09\linewidth}
    \includegraphics[width=\linewidth]{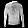}
  \end{subfigure}
  \begin{subfigure}{0.09\linewidth}
    \includegraphics[width=\linewidth]{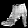}
  \end{subfigure}
  \begin{subfigure}{0.09\linewidth}
    \includegraphics[width=\linewidth]{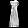}
  \end{subfigure}
  \begin{subfigure}{0.09\linewidth}
    \includegraphics[width=\linewidth]{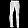}
  \end{subfigure}
  \begin{subfigure}{0.09\linewidth}
    \includegraphics[width=\linewidth]{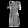}
  \end{subfigure}
  \begin{subfigure}{0.09\linewidth}
    \includegraphics[width=\linewidth]{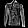}
  \end{subfigure}

    \vspace{0.2em}

    \begin{subfigure}{0.09\linewidth}
    \includegraphics[width=\linewidth]{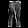}
  \end{subfigure}
  \begin{subfigure}{0.09\linewidth}
    \includegraphics[width=\linewidth]{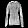}
  \end{subfigure}
  \begin{subfigure}{0.09\linewidth}
    \includegraphics[width=\linewidth]{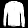}
  \end{subfigure}
  \begin{subfigure}{0.09\linewidth}
    \includegraphics[width=\linewidth]{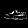}
  \end{subfigure}
  \begin{subfigure}{0.09\linewidth}
    \includegraphics[width=\linewidth]{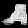}
  \end{subfigure}
  \begin{subfigure}{0.09\linewidth}
    \includegraphics[width=\linewidth]{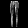}
  \end{subfigure}
  \begin{subfigure}{0.09\linewidth}
    \includegraphics[width=\linewidth]{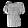}
  \end{subfigure}
  \begin{subfigure}{0.09\linewidth}
    \includegraphics[width=\linewidth]{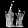}
  \end{subfigure}
  \begin{subfigure}{0.09\linewidth}
    \includegraphics[width=\linewidth]{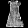}
  \end{subfigure}
  \begin{subfigure}{0.09\linewidth}
    \includegraphics[width=\linewidth]{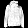}
  \end{subfigure}

  \caption{Fashion-MNIST samples generated with the baseline model (first row) and \algo{} models trained using the \textsc{Full} (second row), \textsc{USplit} (third row), \textsc{ULatDec} (fourth row), and \textsc{UDec} (fifth row) methods with $K = 5$, $R = 15$ and $E = 5$.}
 \label{fig:clothes}
\end{figure}

Lastly, we can see that for $K \in \{2,5\}$, the mean FID scores are below the image artifact threshold for each of the methods. Together with the actual outputs shown in Figure \ref{fig:clothes}, this proves that even with a 74\% reduction in $N$, images with quality comparable to the centralized baseline can be generated in a federated setting with IID data. \textsc{Full} and \textsc{USplit} are also able to deal with $K = 10$, although the FID scores are significantly worse than for $K = 2$ and $K = 5$. \textsc{UDec} and \textsc{ULatDec} fail to produce images of sufficient quality with $K = 10$.  

In general, the FID scores tend to rise as the number of clients increases, suggesting the need to increase either $R$ or $E$ in scenarios involving a larger number of clients. In future work, we therefore plan to plot the FID scores over different higher round numbers, which will require more time than currently available.

\begin{table}[t]
\centering
\caption{FID scores and number of communicated parameters $N$ for different training methods, numbers of clients $K$ and data distributions, using $R=15$ and $E=5$. The baseline uses $E=1$. The * denotes that the FID scores have been averaged over all local client models. Scores that exceed the artifact threshold of 72 within one standard deviation are marked in orange. }
\label{tab:results}
\footnotesize
\begin{tabular}{lrrccc}
\hline
\multirow{2}{*}{\textbf{Method}} &
\multirow{2}{*}{\textbf{K}} &
\multirow{2}{*}{\textbf{N ($\cdot 10^6$)}} &
\multicolumn{3}{c}{\textbf{FID}} \\ \cline{4-6}
 & & & \textbf{IID} & \textbf{l-skew} & \textbf{q-skew}\\ \hline

\hline
 {\textsc{Baseline}} & 1 & 0 & $43 \pm 1$ & n/a & n/a\\
\hline
\multirow{4}{*}{\textsc{Full}}
 & 2  & 179.78 & $39\pm2$ & $33\pm1$  & $33\pm3$\\
 & 5  & 449.45 & $39\pm4$ & $43\pm4$  & $23\pm5$\\
 & 10 & 898.89 & $61\pm2$ & $64\pm3$  & \textcolor{TuOrange}{$76\pm11$}\\ 
 \hline
\multirow{4}{*}{\textsc{USplit}} 
 & 2  & 134.83  & $37\pm3$ & $38\pm4$  & $55\pm4$\\
 & 5  & 343.73  & $41\pm5$ & $61\pm5$  & $39\pm9$\\
 & 10 & 674.17  & $62\pm3$ & \textcolor{TuOrange}{$70\pm8$}  & \textcolor{TuOrange}{$87\pm19$}\\ 
 \hline
\multirow{4}{*}{\textsc{ULatDec*}} 
 & 2  & 105.50 & $45\pm13$ & $49\pm4$ & \textcolor{TuOrange}{$54\pm24$}\\
 & 5  & 263.75 & $53\pm15$ & \textcolor{TuOrange}{$72\pm30$} & \textcolor{TuOrange}{$122\pm138$}\\
 & 10 & 527.51 & \textcolor{TuOrange}{$70\pm14$} & \textcolor{TuOrange}{$101\pm83$} & \textcolor{TuOrange}{$137\pm125$}\\ 
 \hline 
\multirow{4}{*}{\textsc{UDec*}}
 & 2  & 47.54 & $49\pm16$ & $49\pm5$ & \textcolor{TuOrange}{$78\pm48$}\\
 & 5  & 118.85 & $51\pm15$ & \textcolor{TuOrange}{$75\pm31$} & \textcolor{TuOrange}{$139\pm135$}\\
 & 10 & 237.69 & \textcolor{TuOrange}{$72\pm20$} & \textcolor{TuOrange}{$98\pm67$} & \textcolor{TuOrange}{$147\pm119$}\\ 
 \hline
\end{tabular}
\end{table}

\begin{table}[htbp]
    \centering
    \caption{Averaged FID scores for the local client models resulting from \textsc{UDec} and \textsc{ULatDec} training on IID data with $K=5$.}
    \label{tab:client_fids}
    \begin{tabular}{l|l|l}
         \hline
         \textbf{Local Model} & \textsc{UDec} &  \textsc{ULatDec}\\
         \hline
         Client 0 & 44 & 44\\
         Client 1 & 35 & 36\\
         Client 2 & 46 & 55\\
         Client 3 & 71 & 68\\
         Client 4 & 58 & 60\\
         \hline
    \end{tabular}
\end{table}

\begin{figure}[t]
\begin{center}
\includegraphics[width=0.8\linewidth]{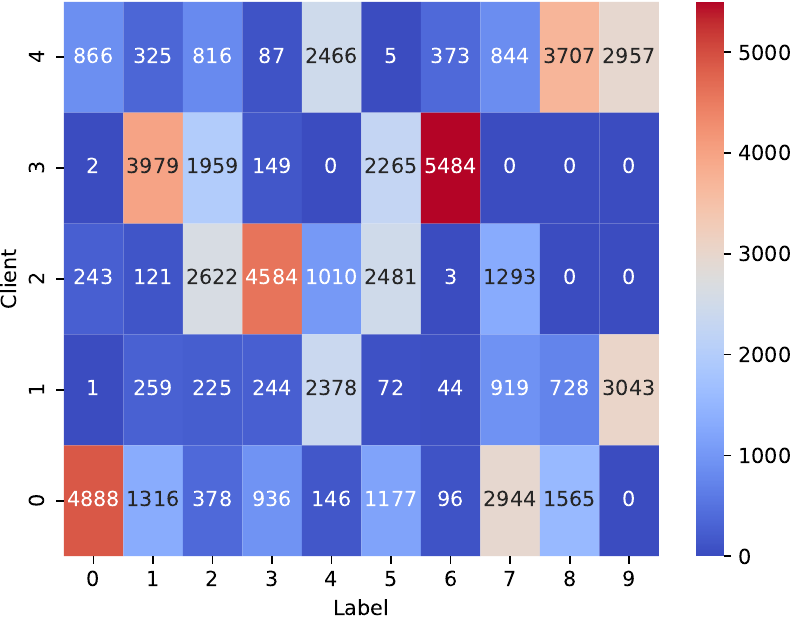} 
\caption{An example of l-skew on the Fashion-MNIST dataset for $K=5$ using $\beta=0.5$. Each cell yields the number of images of a certain label assigned to a certain client.}
\label{fig:skew}
\end{center}
\end{figure}

\textbf{Testing with non-IID data.} To evaluate the robustness of the training methods with respect to statistical heterogeneity, we simulated label distribution skew (l-skew) and quantity skew (q-skew) in our data, using a Dirichlet distribution \cite{dirichlet, dirichlet_2}. To mimic l-skew, we sampled $p_j \sim Dir_K(\beta)$  for every label $j$ and allocated a $p_{j,k}$ proportion of the instances to each client $k$. To mimic q-skew, we sampled $q \sim Dir_K(\beta)$ and allocated a $q_k$ proportion of the total training dataset to each client $k$.
Parameter $\beta$ is the concentration parameter. When $\beta \to \infty$, the result is an IID distribution. The closer $\beta$ is to 0, the more skewed the distribution. We fixed $\beta = 0.5$ as in \cite{dirichlet}. Figure \ref{fig:skew} shows and example of a l-skewed data partition when $K=5$, where every client has a few major classes with many samples, as well as minor classes with relatively few samples. 

\noindent We ran five experiments with l-skew and q-skew for all training methods with $K \in \{2, 5, 10\}$ and reported the averaged FID scores in Table \ref{tab:results}. The combination of a large number of clients $K=10$ together with q-skew or l-skew appeared problematic for most training methods (except \textsc{full}), which is in line with our hypothesis based on GAN results from Section \ref{sec:related_work}. Where \textsc{Full} and \textsc{USplit} were able to cope with q-skew in combination with fewer clients, \textsc{UDec} and \textsc{ULatDec} failed to cope with it at all. Interestingly, \textsc{Full} performed extremely well on q-skewed data with $K=5$, outperforming the IID scenario by far without an explainable reason. \textsc{Full} appeared robust against l-skew, which is in line with findings by \cite{fed_ukd} and resulted in similar FID scores as in the IID setting. However, all other methods seem to be affected by l-skew starting from $K=5$, leading to notable drops in image quality compared to the IID setting. \\

\begin{figure}[t]
  \centering
  \begin{subfigure}{0.18\linewidth}
    \includegraphics[width=\linewidth]{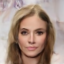}
  \end{subfigure}
  \begin{subfigure}{0.18\linewidth}
    \includegraphics[width=\linewidth]{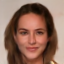}
  \end{subfigure}
  \begin{subfigure}{0.18\linewidth}
    \includegraphics[width=\linewidth]{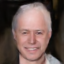}
  \end{subfigure}
  \begin{subfigure}{0.18\linewidth}
    \includegraphics[width=\linewidth]{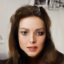}
  \end{subfigure}
  \begin{subfigure}{0.18\linewidth}
    \includegraphics[width=\linewidth]{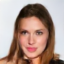}
  \end{subfigure}

  \begin{subfigure}{0.18\linewidth}
    \includegraphics[width=\linewidth]{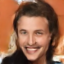}
  \end{subfigure}
  \begin{subfigure}{0.18\linewidth}
    \includegraphics[width=\linewidth]{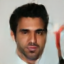}
  \end{subfigure}
  \begin{subfigure}{0.18\linewidth}
    \includegraphics[width=\linewidth]{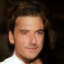}
  \end{subfigure}
  \begin{subfigure}{0.18\linewidth}
    \includegraphics[width=\linewidth]{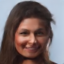}
  \end{subfigure}
  \begin{subfigure}{0.18\linewidth}
    \includegraphics[width=\linewidth]{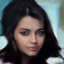}
  \end{subfigure}

  \caption{CelebA samples generated with a \algo{} model trained using the \textsc{Full} method on IID data with $K=5$.}
  \label{fig:faces}
\end{figure}

\textbf{Testing with other Datasets.} The choice for the Fashion-MNIST dataset allowed for fast training and evaluation. However, training diffusion models using low-resolution grayscale images forms a drastic simplification of real-world diffusion training tasks. Hence, we were interested in experimenting with higher dimension colored images too. We chose the CelebA dataset \cite{celeba}, which contains over 200k images of celebrities for this purpose. We resized the images to 64x64 and to facilitate the creation of different data distributions, we created 16 different classes among the images based on the combination of sex (male, female), age (young, old) and hair color (black, brown, blond, gray). As some images were not annotated properly, we ended up with a usable dataset comprising of 162,770 training images and 19,962 test images. Using \algo{}, we trained a 14,892,477 parameter model over an IID dataset with $K = 5$, $R = 30$, $E = 5$ and $B = 64$, which took over 37 hours. We were able to determine a FID score of 53 after a 5 hour sampling process. The federated model demonstrated its ability to generate realistic faces, as depicted in Figure \ref{fig:faces}. 

\section{Conclusion} 
\label{sec:conclusions}

We have demonstrated that diffusion models can be trained using federated learning by utilizing an adapted Federated Averaging (FedAvg) algorithm to train a UNet-based Denoising Diffusion Probabilistic Model (DPPM). Moreover, we have shown that the images generated by our federated model exhibit comparable quality to those generated by their non-federated counterparts, as evaluated by the FID score. We have also shown our method's robustness to label and quantity-skewed data distributions. 

Furthermore, we discovered that additionally  splitting the parameter updates for the encoder, decoder, and bottleneck parts of the UNet among clients every round can enhance communication efficiency during training. This approach led to a 25\% reduction in the number of exchanged parameters whilst maintaining image quality comparable to the naive approach, where all parameters are exchanged between the federator and clients every round. However, this method demonstrated limited resilience against label and quantity skew in a federated setting with few clients.

We also found that training the encoder and bottleneck locally resulted in a significant reduction in communication by up to 74\% compared to the naive approach. However, this approach exhibited variations in image quality among the local client models and was only effective when applied to a limited number of clients in conjunction with IID data.
\bibliographystyle{IEEEtran}
\bibliography{references}
\end{document}